\documentclass[VANCOUVER,STIX1COL]{WileyNJD-v2}

\usepackage{subfigure}

\articletype{Article Type}%

\received{26 April 2019}
\revised{6 June 2019}
\accepted{6 June 2019}

\begin{document}

\title{Real-time Segmentation and Facial Skin Tones Grading\protect\thanks{Part of this work was done when L. Luo was an intern at Meidaojia, Beijing, China.}}

\author[1]{Ling Luo*}

\author[1]{Dingyu Xue}

\author[1]{Xinglong Feng}

\author[2]{Yichun Yu}

\author[3]{Peng Wang}

\authormark{L. Luo \textsc{et al}}

\address[1]{\orgname{Northeastern University}, \orgaddress{\state{Shenyang}, \country{China}}}

\address[2]{\orgname{Beihang University}, \orgaddress{\state{Beijing}, \country{China}}}

\address[3]{\orgname{Microsoft Research Asia}, \orgaddress{\state{Beijing}, \country{China}}}

\corres{*Ling Luo, Northeastern University. \email{lingluo@stu.neu.edu.cn}}

\presentaddress{Meidaojia, Beijing, China}

\abstract[Summary]{Modern approaches for semantic segmention usually pay too much attention to the accuracy of the model, and therefore it is strongly 
recommended to introduce cumbersome backbones, which brings heavy computation burden and memory footprint. To alleviate this problem, we propose an 
efficient segmentation method based on deep convolutional neural networks (DCNNs) for the task of hair and facial skin
segmentation, which achieving remarkable trade-off between speed and performance on three benchmark 
datasets. As far as we know, the accuracy of skin tones classification is usually unsatisfactory due to the influence of external environmental 
factors such as illumination and background noise. Therefore, we use the segmentated face
to obtain a specific face area, and further exploit the color moment algorithm to extract its color features.
Specifically, for a 224\ $\times$\ 224 standard input, using our high-resolution spatial detail information
and low-resolution contextual information fusion network (HLNet), we achieve \textbf{90.73}\% Pixel
Accuracy on Figaro1k dataset at over \textbf{16} FPS in the case of CPU environment. Additional experiments
on CamVid dataset further confirm the universality of the proposed model. We further use masked
color moment for skin tones grade evaluation and approximate \textbf{80}\% classification accuracy
demonstrate the feasibility of the proposed scheme. Code is available at \url{https://github.com/JACKYLUO1991/Face-skin-hair-segmentaiton-and-skin-color-evaluation}.}


\jnlcitation{\cname{%
        \author{L. Luo},
        \author{D. Xue},
        \author{X. Feng},
        \author{Y. Yu}, and
        \author{P, Wang}} (\cyear{2019}),
    \ctitle{Real-time Segmentation and Facial Skin Tones Grading}, \cjournal{Computer Animation \& Virtual Worlds}, \cvol{2019;12:1--6}.}

\maketitle

\section{Introduction}\label{sec1}
AR (Augmented Reality) technology has become a hot spot recently due to its widespread
application in various domains, and the most widely used is the beauty industry.
Among them, automatic hair dyeing (in Fig. 1) is one of the major applications in the beauty
industry. However, there are enormous challenges in the actual application scenario.
First of all, because the hair has a very complex shape structure \cite{muhammad2018}, it is quite difficult
to handle accurate edge information. Although the existing semantic segmentation
methods\cite{Long2014}\cite{Ronneberger2015}\cite{Yu2015} have relatively high segmentation performance for simple objects, only a
relatively rough mask can be obtained in the processing of hair segmentation task.
Secondly, almost all networks require GPU with high computation power that most of
mobile devices do not have. It greatly limits the using scenario. Thirdly, taking into account 
runtime limitations, Conditional Markov random fields (CRFs) \cite{Zheng2015} is not suitable for edge processing. 
Taking all these factors into considerations, real-time hair dyeing
faces enormous challenges. At the same time, e-commerce and digital interaction with
the clients allows people to buy their favorite products without leaving home. Among
them, the robust product recommendation function plays an important role. Automatic
assessment of skin tones levels which makes it possible to personalize recommendation
for beauty products. However, in the context of a complex environment, skin tones
grading is influenced by lighting, shadows, and imaging equipment etc. that even an
experienced skin therapist is difficult to judge with the naked eye. This paper is
dedicated to solving these problems using machine learning and fiery deep learning
algorithms. 

Semantic segmentation is an advanced visual task, whose goal is to assign different category labels 
to each pixel. However, limited by bulky backbones, existing state-of-the-art models are not suitable for actual 
deployment. In this paper, we strive to balance the relationship between the efficiency and speed of
segmentation network, and provide a much simpler and more compact alternative for
our multi-task segmentation scenario. In order to get accurate segmentation results, global 
information and context information should be considered simultaneously. 
Based on this observation, we propose a spatial and context information fusion framework
(HLNet) that high-dimensional and low-dimensional feature maps are integrated into
one network. Further experiments confirm that HLNet achieve significant trade-off between efficiency and accuracy. 
Considering that background illumination is not conducive to identifying skin tones,
we extract features (\emph{a.k.a.} masked color moments) based on the segmented face and color moment algorithm.
After that, the mask color moments are input into a powerful Random Forest Classifier to evaluate 
a person's skin tone level.

The remainder of the paper is organized as follows. In Section 2, we briefly
review the latest real-time and efficient deep-learning based semantic segmentation
proposals as well as domain-specific segmentation algorithms in the field of computer
vision. We elaborate on the process of our scheme in Section 3, then ablation and 
contrast experiments are performed in Section 4. At last, we summarize our work and
future research directions in Section 5.

\section{Related Work}\label{sec2}

\begin{figure}[t]
\centering
\subfigure[]{
    \includegraphics[height=5cm]{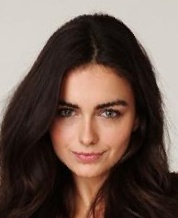}
}
\subfigure[]{
    \includegraphics[height=5cm]{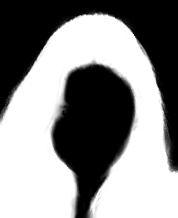}
} 
\subfigure[]{
    \includegraphics[height=5cm]{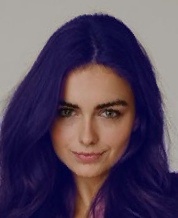}
}  
\caption{Automatic hair dyeing exemplar. (a) Input RGB image. (b) The guided filter output of our proposed algorithm. (c) Final dyed rendering.\label{fig1}}
\end{figure}

\subsection{Lightweight model}
Since pioneering work \cite{Long2014} based on deep learning, many high-quality backbones\cite{Chen2017}\cite{Jegou2017}\cite{Zhao2017}\cite{Lin2017}
have been derived. However, due to the requirements of computationally limited
platforms (e.g., drones, autonomous driving, smartphone), people pay more attention to the efficiency of the network than just the
performance.

\emph{ENet}\cite{Paszke2016} is the first lightweight network for real-time scene segmentation which does not
apply any post-processing steps in an end-to-end manner. Zhao \emph{et al}.\cite{Zhao2018} introduces a
cascade feature fusion unit to quickly achieve high-quality segmentation. Howard \emph{et
al}.\cite{Howard2017} present a compact encoder module which based on a streamlined architecture
that uses Depthwise separable convolutions to build light-weight deep neural
networks. Poudel \emph{et al}.\cite{Poudel2019} combine spatial detail at high resolution with deep features
extracted at lower resolution yielding beyond real-time effects. Recently, \emph{LEDNet}\cite{Wang2019}
has been proposed which channel split and shuffle are utilized in each residual block
to greatly reduce computation cost while maintaining higher segmentation accuracy.

\subsection{Contextual information}
Some details cannot be recovered during conventional up-sampling of the feature
maps to restore the original image size. The design of \emph{skip connections}\cite{He2016} can alleviate
this deficiency to some extent. Zhao \emph{et al}.\cite{Zhao2017} propose a pyramid pooling module that
can aggregate context information from different regions to improve the ability to
capture multi-scale information. 
Zhang \emph{et al}.\cite{Zhang2018} design a context encoding module to
introduce global contextual information, which is used to capture the context
semantics of the scene and selectively highlight the feature map associated with a
particular category. Fu \emph{et al}.\cite{Fu2019} address the scene parsing task by capturing rich contextual
dependencies based on spatial and channel attention mechanisms, which significantly improve 
the performance on numerous challenging datasets.

\subsection{Post processing}
Generally, the quality of the above segmentation methods is obviously rough and requires additional post-processing operations. 
Post-processing mechanisms are usually able to improve image
edge detail and texture fidelity, while maintaining a high degree of
consistency with global information. Chen \emph{at el}.\cite{Chen2014} propose a CRF post processing
method that overcome poor localization in a non-end-to-end way. \emph{CRFasRNN}\cite{Zheng2015}
considers the CRF iterative reasoning process as an RNN operation in an end-to-end
manner. In order to eliminate the excessive execution time of CRF, Levinshtein \emph{at al}.\cite{Levinshtein2018}
present a hair matting method with real-time performance on mobile devices

\subsection{Color feature extraction}

\emph{Color Histograms}\cite{Swain1991} are widely used color features in many image retrieval
systems which describe the proportion of different colors in the entire image.
Calculating the color histogram requires dividing the color space into a number of
small color intervals, each called a bin. This process is often referred to as color
quantization. Since the color space such as HSV and LAB is more in line with
people's subjective judgment on color similarity, RGB space is not generally used.
The biggest disadvantage of this method is that it cannot represent the local
distribution of colors in the image and the spatial location of each color.

\emph{Color Moment}\cite{Stricker1995} proposed by Stricker and Orengo is another straightforward and
effective color feature. The mathematical basis of this method is that any color
distribution in an image can be represented by its moment. In addition, since the color
distribution information is mainly concentrated in the low-order moment, only the
first moment (mean), the second-order moment (variance) and third-order moment
(skewness) of the color are sufficient to express the color distribution of the image. 
Another benefit of this approach is that it does not require vectorization of features
compared to color histograms. 

The \emph{Color Correlogram}\cite{Huang1997} is also an expression of the color distribution of the
images. This feature not only depicts the proportion of pixels in a certain color to the
entire image, but also reflects the spatial correlation between pairs of different colors.
However, this solution is too complicated in time. 

Overall, our approach is closely related to asymmetric encoding and decoding structure.
Furthermore, we simply employ masked color moment to sort the skin tones, 
which will be discussed in the next chapter.

\begin{figure}[h]
    \centerline{\includegraphics[width=\textwidth]{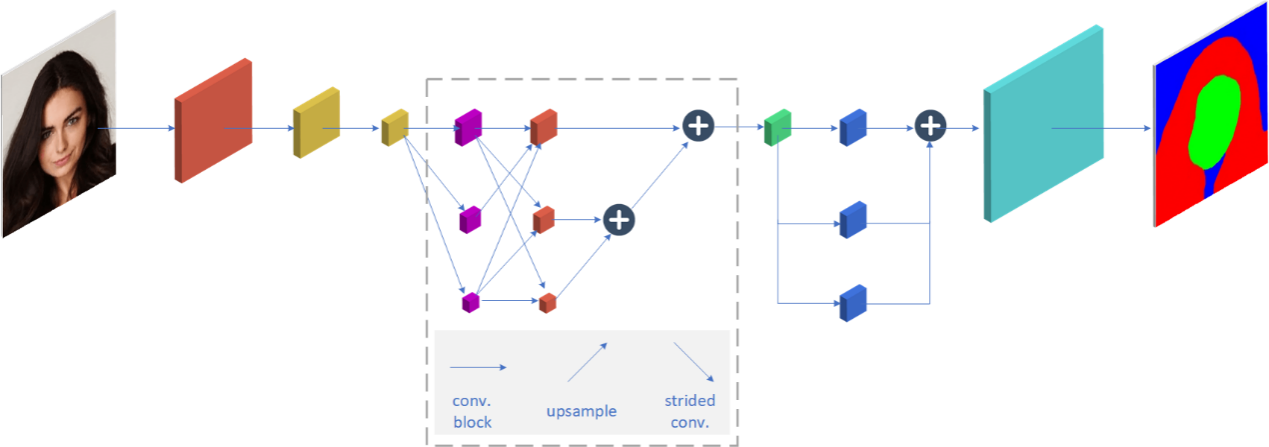}}
    \caption{An overview of our asymmetric encoder-decoder network. Blue, red and green
    represents background, hair mask recolored and face mask recolored, respectively. In the dotted rectangle (also
    called \emph{InteractionModule}), different arrows represent different operations. `+' means \emph{Add} operation.\label{fig2}}
\end{figure}

\section{Methodology}\label{sec3}
In this section, we will elaborate the proposed method and the model framework 
for hair and face segmentation, as well as the methodological basis for 
subsequent facial skin tones classification. 

\subsection{High-to-low dimension fusion network}
The proposed HLNet network is inspired by \emph{HRNet}\cite{Sun2019} which maintains high-resolution
representation through the whole process by connecting high-to-low resolution
convolutions in parallel. Figure 2 illustrates the overall framework of our model. We experimentally
prune the model parameters to increase the speed without excessive performance
degradation. Furthermore, the existing state-of-the-art modules\cite{Poudel2019}\cite{Sandler2018}\cite{Yu2018}\cite{Chollet2017} are
reasonably combined to further improve the performance of the network.

\textbf{Architecture.} Table 1 gives an overall description of the modules involved in the
network. The model consists of different kinds of convolution modules, up-sampling (\textbf{complementary}: transposed convolution layer can cause
gridding artifacts), bottlenecks, and other feature maps communication modules. In the following part,
we will expand the above modules in detail.

In the first three layers, we refer to \emph{Fast-SCNN}\cite{Poudel2019} to employ standard convolution and depth
separable convolution for fast down-sampling in order to ensure low-level feature
sharing. Depth separable convolution reduces the amount of model parameters
effectively while achieving a comparable representation ability. The above convolution
unified use stride 2 and 3\ $\times$\ 3 spatial kernel size, followed by BN\cite{Ioffe2015} and RELU
activation function. According to \emph{FCOS}\cite{Tian2019}, the low-dimensional detail information of the feature map
promotes the segmentation of small objects, so we superimpose the low-dimensional
layers to enhance detail representation ability of the proposed model. Interaction of
high-resolution and low-resolution information facilitates learning of multi-level
information representation. In this part, we draw the above advantages and propose a information interaction 
module (InteractionModule) with feature maps of different resolutions to obtain elegant output results. 
Different feature map scales use 1\ $\times$\ 1 convolution and
Bilinear up-sampling module for channel combination and exchange. Up-sampling is
characterized by the fact that no weight parameters need to be learned, so we discard
the operation of transposed convolution to save computing resources. In addition to 
standard convolution modules, we also use the efficient inverted bottleneck residual 
block modules proposed by \emph{MobileNet v2}\cite{Sandler2018} which is illustrated in Figure 3. Subsequently, we followed the \emph{FFM Attention}\cite{Yu2018} 
to focus the model on the channel features with the most information and suppress those channel features that are not important.
To cature multi-scale context information, we also introduce a multi-receptive field fusion block (\emph{e.g}. dilation rate is set to 2, 4, 8).

For simplicity, the decoder directly performs bilinear upsampling on the 28\ $\times$\ 28 feature map (\textbf{Complementary}: transposed convolution 
layer can cause gridding artifacts\cite{Levinshtein2018}), and additionally adds a \emph{SOFTMAX} layer for 
pixel-wise classification.

\begin{figure}[h]
    \centerline{\includegraphics[height=2.5cm]{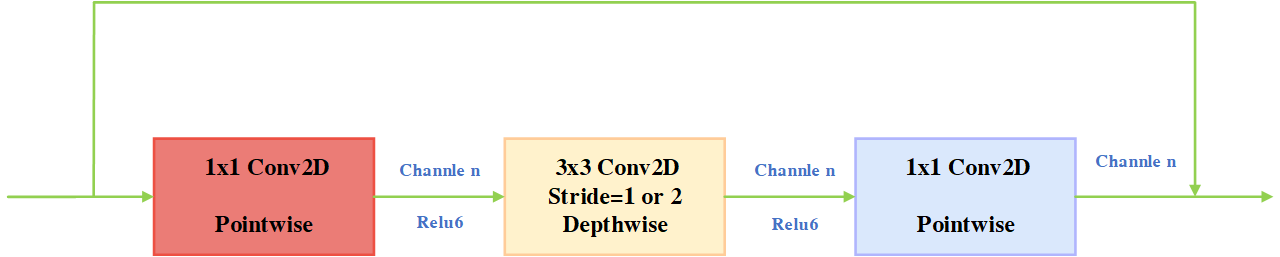}}
    \caption{Inverted residual module proposed by \emph{MobileNet v2} release. If the stride size is equal to 1, additional
    shortcut operation is required. In general, t=6 enables the separation convolution to
    extract features in a higher dimension, which effectively improves the representation ability of the model.\label{fig3}}
\end{figure}

\textbf{Post Processing.} In pursuit of perceptual consistency and reduce the time complexity
of running, we advocate the idea of \emph{Guided Filter}\cite{He2010}\cite{He2015} to achieve edge-preserving and
denoising. \emph{Guided Filter} can effectively suppress gradient-reversal artifacts and
produce visually pleasing edge profiles. Given a guidance image $I$ and filtering
input image $P$, our goal is to learn a local linear model to describe the relationship
between the former and the output image $Q$ while seeking consistency between $P$ and
$Q$ just like the role of \emph{Image Matting}\cite{Levin2007} For simplicity and efficiency, $\mathit{s, r, \zeta}$
are set to $4$, $4$, and $50$, respectively, during the experiment.

\subsection{Facial skin tones classification}
The second stage is to classify facial skin tones. Usually for Asians, we divide it into
\emph{porcelain white, ivory white, medium, yellowish} and \emph{black}. For skin tones features, it
is not wise to use deep learning for feature extraction due to its feature space is
relatively small, which is easy to cause under-fitting. Therefore, after repeated
thoughts and experimental trial and error, the scheme is selected to extract the color
moment of the image as the features to be learned and put it into the classic machine
learning algorithm for learning. Considering facial skin tones in complex scenes, background lighting has 
a incurable impact on the results. So we employ image morphology algorithms and pixel-level operations to 
get rid of background interference. Ultimately, the remaining face regions are used to
extract the color moment features and then fed into the machine learning algorithm
for learning. As far as the classifier is concerned, we choose the \emph{Random Forest
Classifier}\cite{Pal2005} due to its powerful classification ability. The detail of algorithm is summarized in \textbf{Algorithm 1}.

\begin{algorithm}
    \caption{\textbf{:} Framework of facial region extraction.}\label{alg1}
    \begin{algorithmic}
        \item \textbf{Input:} Original rgb image $I_{i}$, totol number of images $N$.
        \item \textbf{Ouput:} Smooth facial region $F_{i}$ after image processing.
        \item \textbf{Initial stage:} Get the segmentation mask $M_{i}^c(c\subseteq0,1,2)$ with the help of our \emph{HLNet}.
        \For{each $i\in N$}
            \State Extract rough facial mask region $M_{i}^2$ (For the convenience of description, it 
            is referred to as $M_{i}$), 
            \State Referring to \emph{Image Erode Formula} \textbf{dst} ${P_{i}(x, y)}$ $\longleftarrow$ min \textbf{src} ${M_{i}(x + x^{'}, y + y^{'})}$ 
            \textbf{where} element$(x^{'}, y^{'})\not=0$ in order to 
            \State eliminate mask gaps,
            \State Use \emph{Bilateral Filtering} to smooth mask edges, as $Q_{i}$,
            \State The final $F_{i}$ is obtained by \emph{Bitwise and} operation between $I_{i}$ and $Q_{i}$.
        \EndFor
    \end{algorithmic}
\end{algorithm}

\section{Experimental Evaluation}\label{sec4}
we evaluate the segmentation performance 
of hair and skin segmentation on three public datasets and further confirm the generalization of the proposed
architecture on the challenging benchmark vehicle road dataset \emph{CamVid}\cite{Brostow2008}. In the
following, we begin with a brief introduction to the datasets which involved in
the experiment, including a manually annotated dataset. Next, by comparing with the
existing scheme and conducting ablation experiment, it is proved that our scheme
could achieve outstanding trade-off between speed and accuracy. All the networks mentioned below 
follow the same training strategy, except for the
initial learning rate setting. They are trained using mini-batch stochastic gradient
descent (SGD) with batch size $\mathrm{64}$, momentum $\mathrm{0.98}$ and weight decay 2e-5 for face-related segmentation datasets. 
For \emph{CamVid} experiment, we use
Adam gradient descent strategy with learning rate 1e-3 because we found it more
conducive to the convergence of the model. The former adopts `poly' learning rate policy in configuration where the initial rate is multiplied by $(1-\frac{iter}{total\_iter})^{power}$ with power $\mathrm{0.9}$ and 
initial learning rate is set as 2.5e-3. In terms of loss function, we
apply \emph{generalized dice loss}\cite{Sudre2017} to compensate for the segmentation performance of
small objects. Data augmentation includes normalization, random rotation $\mathrm{[-20, 20]}$,
random scale $\mathrm{[-20, 20]}$, random horizontal flip and random shift $\mathrm{[-10, 10]}$. For fair comparison, all the methods are
 conducted on a server equipped with a single NVIDIA GeForce GTX$\mathrm{1080}$Ti GPU.

\subsection{Database introduction}
Data is the soul of deep learning, because it determines the upper limit of an
algorithm to some extent. In order to ensure the robustness of the algorithm, it is
necessary to construct a dataset with human faces in extreme situation such as large
angles, strong occlusions, complex lighting changes etc.

\subsubsection{Face and hair segmentation datasets}
\emph{Labeled Faces in the wild} (LFW)\cite{Kae2013} dataset consists of more than 13,000 images
on the Internet. We use its extension version (Part Labels) during the experiment
which automatically labeled via a super-pixel segmentation algorithm. We adopt the
same data partitioning method in\cite{Borza2018} as 1500 images in the training, 500 used in
validation, and 927 used for testing.

\emph{Large-scale CelebFaces Attributes dataset} (CelebA)\cite{Yang2015} consisting of more than 200k
celebrity images, each with multiple attributes. The main advantage of this dataset
is that it combines large pose variations and background clutter making the
knowledge learned from this dataset easier to satisfy demand of actual products. In the
experiment we adopted the version of CelebA dataset in\cite{Borza2018} which includes 3556
images. We used the same configuration parameters as the original paper, that is, 20$\%$ 
for validation.

The last dataset of the experiment, we employ \emph{Figaro1k}\cite{Svanera2016} dedicated to hair
analysis. However, as this dataset was developed for general hair detection, face is not 
included in many images. In this case, we apply a universal face detector as \cite{muhammad2018}, which was
proposed by the literature. Since only the image of the detected face can
participate in subsequent segmentation tasks, 171 images are used for experiments.

\subsubsection{Benchmark CamVid dataset}
In order to demonstrate the generalization ability of the proposed method, we also
conduct experiments on benchmark CamVid dataset. \emph{CamVid} dataset contains images
extracted from video sequences with resolution up to 960\ $\times$\ 720. In this experiment, we
adopt the same setting as\cite{Sturgess2009} which contains 701 images in total, in which 367 for
training, 101 for validation and 233 for testing.

\begin{figure}[t]
    \centering
    \subfigure[]{
        \includegraphics[height=4cm]{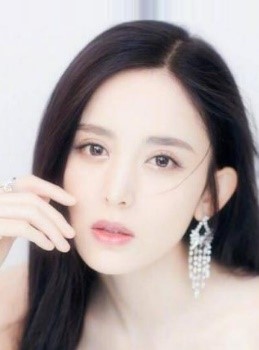}
    }
    \subfigure[]{
        \includegraphics[height=4cm]{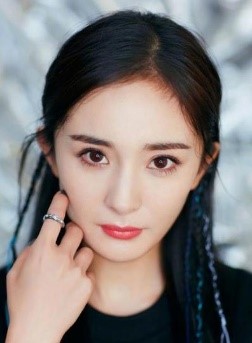}
    } 
    \subfigure[]{
        \includegraphics[height=4cm]{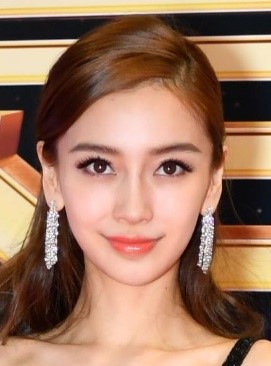}
    }
    \subfigure[]{
        \includegraphics[height=4cm]{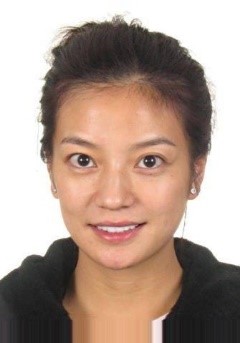}
    } 
    \subfigure[]{
        \includegraphics[height=4cm]{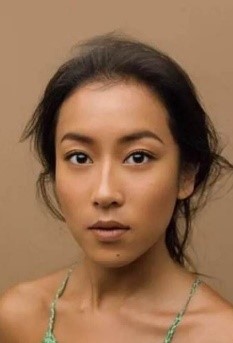}
    } 
\caption{Manually marked face skin hue level sample after voting mechanism. From left to right, it represents
porcelain white, ivory white, medium, yellowish and black. In order to have a consistent understanding of image
classification criteria, only actresses are used here.\label{fig4}}
\end{figure}

\subsubsection{Annotated dataset}
One contribution of this work is a manually labeled face skin tones grading dataset.
According to different facial tones hue levels of Asians, it is mainly divided into five
categories: porcelain white, ivory white, medium, yellowish and black. In the process
of labeling, three professionally trained makeup artists rated the face tones color using
a voting mechanism. Our face data is also derived from the Internet, leaving only data
that contains faces. These data are in turn used for later feature extraction and
machine learning. The number of each category is 95, 95, 96, 93 and 94, samples are
shown in Figure 4.

\begin{figure}[ht]
    \centering
    \begin{minipage}{0.93\linewidth}
        \centering
        \subfigure[RGB image]{
            \includegraphics[width=3cm]{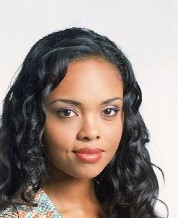}\hspace{2pt}
            \includegraphics[width=3cm]{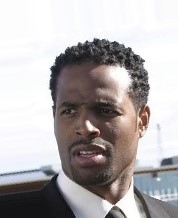}\hspace{2pt}
            \includegraphics[width=3cm]{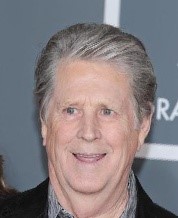}\hspace{2pt}
            \includegraphics[width=3cm]{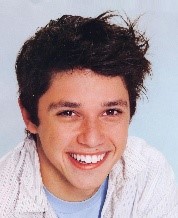}\hspace{2pt}
            \includegraphics[width=3cm]{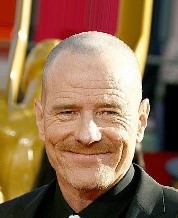}
        }
        \subfigure[Ground truth]{
            \includegraphics[width=3cm]{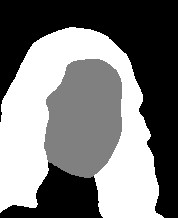}\hspace{2pt}
            \includegraphics[width=3cm]{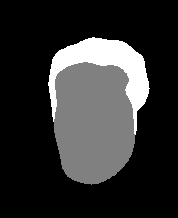}\hspace{2pt}
            \includegraphics[width=3cm]{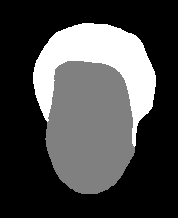}\hspace{2pt}
            \includegraphics[width=3cm]{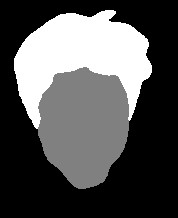}\hspace{2pt}
            \includegraphics[width=3cm]{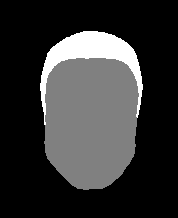}
        }
        \subfigure[Hair output]{
            \includegraphics[width=3cm]{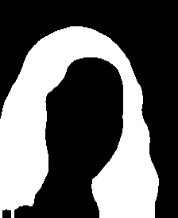}\hspace{2pt}
            \includegraphics[width=3cm]{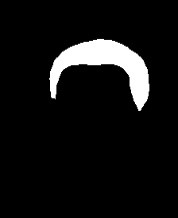}\hspace{2pt}
            \includegraphics[width=3cm]{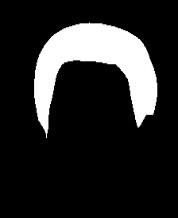}\hspace{2pt}
            \includegraphics[width=3cm]{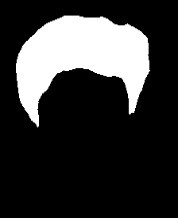}\hspace{2pt}
            \includegraphics[width=3cm]{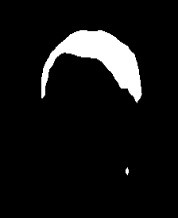}
        } 
        \subfigure[Guided output]{
            \includegraphics[width=3cm]{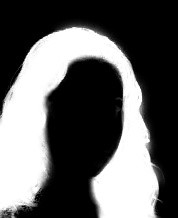}\hspace{2pt}
            \includegraphics[width=3cm]{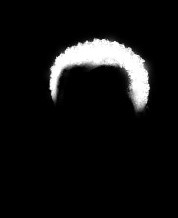}\hspace{2pt}
            \includegraphics[width=3cm]{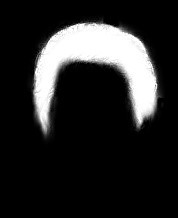}\hspace{2pt}
            \includegraphics[width=3cm]{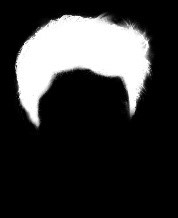}\hspace{2pt}
            \includegraphics[width=3cm]{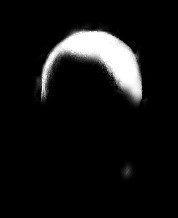}
        }
        \subfigure[ Hair dying]{
            \includegraphics[width=3cm]{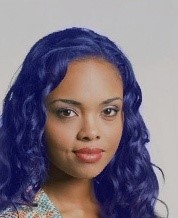}\hspace{2pt}
            \includegraphics[width=3cm]{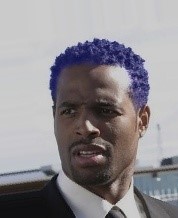}\hspace{2pt}
            \includegraphics[width=3cm]{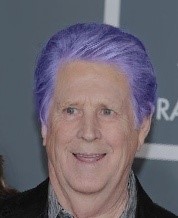}\hspace{2pt}
            \includegraphics[width=3cm]{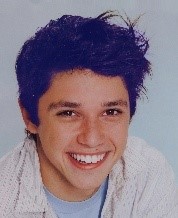}\hspace{2pt}
            \includegraphics[width=3cm]{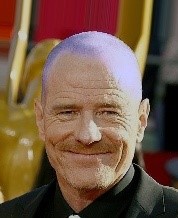}
        }
    \end{minipage}
    \caption{Samples of hair segmentations. From top to the bottom: RGB image, Ground truth, Hair output, Guided output and Hair dying.
    The last column gives the wrong case, which also plagues the human marker.\label{fig5}}
\end{figure}

\subsection{Segmentation results}
In the testing phase, we first use the existing \emph{MTCNN}\cite{Zhang2016} with high recall rate to
extract the face ROI. Secondly, considering the redundant environmental information
has a certain promotion effect on the segmentation background, the ROI region is
enlarged with a factor of 0.8 in both horizontal and vertical dimensions. For
quantitative evaluation, we use four \emph{FCN}\cite{Long2014} derived metrics to evaluate the
performance of hair and face segmentation algorithms. Table 2 shows the comparison results between our method 
and the methods in literature.\cite{Borza2018}

One drawback of fast downsampling is that the feature extraction for the shallow
layer is not sufficient, so our HLNet is slightly worse than VGG-like networks in
CelebA dataset (CelebA facial details are significantly clearer than LFW). However,
considering the running time, we reach 63 ms per image in the CPU terminal without
any tricks. We can further reach no more than 5 ms under GPU. Comparing VGG
with HLNet (64 ms vs 4.7$\pm$0.2 ms) shows that the latter is more efficient, while
performance is comparable. This conclusion suggests that we can further apply this
framework to the edge and embedded devices with small memory and battery budget.
The qualitative analysis results are shown in Figure 5. The post-processing employs
\emph{Guided Filter} to achieve more realistic edge effects.

\textbf{Ablation Study} We conduct the ablation experiments on Figaro1k test dataset, 
and follow the same training strategy for the fairness of the experiments. 
We mainly evaluate the impact of InteractionModule (IM) and DilatedGroup (DG) components on the results, as shown in Table 3.
Three parallel 3 $\times$ 3 convolutions and convolutions with a rate of 1 are used to replace the corresponding components as \emph{baseline}.
When we append DG and IM modules respectively, mIoU increases by 1.54\% and 3.19\% relative to the \emph{baseline}.
When we apply two modules at the same time, mIoU increases dramatically by 4.26\%. The obvious performance gains illustrate
the efficiency of the proposed model.

We also give a comparison of our method with the existing methods on the CamVid dataset. We adopt widely used
mean-interesction-over-union (mIoU) to evaluate segmentation quality. The results
are reported in Table 4. Our HLNet get much faster inference speed while achieving
comparable performance compared to the state-of-the-art models.

\subsection{Facial skin tones classification results}
In the second phase of the experiment, we use ablation study to compare the
influence of different color spaces and different experimental protocols on the results.

\begin{figure}[htbp]
    \centerline{\includegraphics[height=8cm]{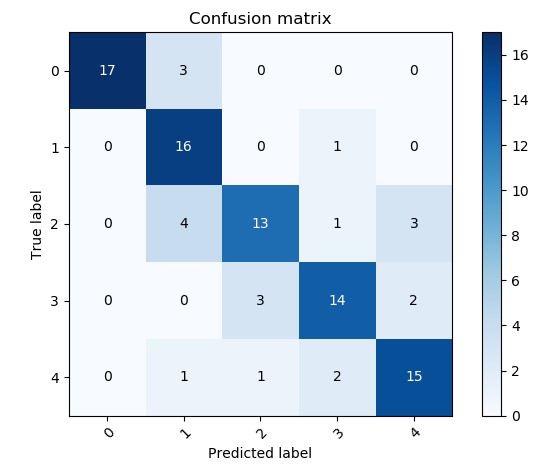}}
    \caption{Multi-classification confusion matrix.\label{fig6}}
\end{figure}

In Table 5, we report the facial skin tones classification accuracy. The best results
are obtained using the YCrCb space with color moment backend. It should be noted
that before the experiment, the data is first oversampled to ensure that the imbalance
between the categories is eliminated. We simply split the dataset into 8:2 for training
and testing, and then use the powerful Random forest classifier for training. Figure 6
gives the confusion matrix for this configuration. It can be observed from the figure
that the main mistakes are between adjacent categories, and this situation also plagues
a trained professional makeup artist when he or she labels data.

\section{Conclusion}\label{sec5}
In this paper, we propose a fully convolutional network to solve the real-time
segmantic segmentation problem, so as to achieve a trade-off between speed and
performance. Through comparative experiments and cross-dataset evaluation, we
prove the feasibility and generalization of our method. Next, we propose a method for
extracting skin tones features, which extracts masked facial tones features and throws
them into a random forest classifier for classification. 80$\%$ classification accuracy
demonstrate the effectiveness of the proposed solution.

The purpose of this work is to apply our algorithm to real-time dyeing, face swap,
skin tones rating system, and skin care product recommendation based on skin tones
levels in realistic scenarios. As a future work, we plan to further explore color features
to improve classification accuracy.

\section*{Acknowledgments}
This work has been supported by the Project of the National Natural 
Science Foundation of China No.61773104.

\subsection*{Conflict of interest}
No potential conflict of interests.

\nocite{*}
\bibliography{wileyNJD-Vancouver}%

\begin{center}
    \begin{table}[h]%
        \centering
        \caption{ HLNet consists of an asymmetric encoder and decoder. The whole network is mainly composed of
        standard convolution (Conv2D), deep separable convolution (DwConv2D), inverted residual bottleneck blocks,
        upsampling (UpSample2D) module and specially designed modules.\label{tab1}}%
        \begin{tabular*}{240pt}{@{\extracolsep\fill}clc@{\extracolsep\fill}}%
        \toprule
        \textbf{Satge} & \textbf{Type} & \textbf{Output Size} \\
        \midrule
        \multirow{7}{*}{\rotatebox{90}{\textbf{Encoder}}} & - & 224 $\times$ 224 $\times$ 3\\
         & Conv2D & 112 $\times$ 112 $\times$ 32 \\
         & DwConv2D & 56 $\times$ 56 $\times$ 64\\
         & DwConv2D & 28 $\times$ 28 $\times$ 64\\
         & InteractionModule & 28 $\times$ 28 $\times$ 64\\
         & FFM\cite{Yu2018} & 28 $\times$ 28 $\times$ 64\\
         & DilatedGroup & 28 $\times$ 28 $\times$ 32\\
        \hline
        \multirow{3}{*}{\rotatebox{90}{\textbf{Decoder}}} 
        & UpSample2D & 224 $\times$ 224 $\times$ 32\\
        & Conv2D & 224 $\times$ 224 $\times$ 3\\
        & SoftMax & 224 $\times$ 224 $\times$ 3\\
        \bottomrule
        \end{tabular*}
    \end{table}
\end{center}

\begin{center}
    \begin{table*}[h]%
    \caption{Segmentation performance of LFW, CelebA and Figaro1k dataset. Comparative analysis shows that our
    network can still achieve wonderful results when the parameter quantity is much smaller than the other two, in
    some metrics, even exceed the cumbersome VGG\cite{Long2014} network. All values are in \%.\label{tab2}}
    \centering
    \begin{tabular}{l|lll|lll|lll}
    \hline
    \multirow{2}{*}{} & \multicolumn{3}{c|}{\textbf{LFW}}       & \multicolumn{3}{c|}{\textbf{CelebA}} & \multicolumn{3}{c}{\textbf{Figaro1k}} \\ \cline{2-10} 
                    & VGG            & U-Net & HLNet          & VGG              & U-Net   & HLNet   & VGG      & U-Net   & HLNet            \\ \hline
    mIoU              & 87.09          & 83.46 & \textbf{88.03} & \textbf{92.03}   & 88.56   & 91.90   & 77.79    & 77.75   & \textbf{78.39}   \\
    fwIoU             & \textbf{94.67} & 92.75 & 94.26          & \textbf{94.42}   & 91.79   & 93.65   & 82.66    & 83.01   & \textbf{83.12}   \\
    pixelAcc          & \textbf{97.01} & 95.83 & 96.01          & \textbf{97.06}   & 95.54   & 96.78   & 90.15    & 90.28   & \textbf{90.73}   \\
    mPixelAcc         & \textbf{91.81} & 88.84 & 91.26          & \textbf{95.77}   & 93.61   & 95.53   & 84.75    & 84.72   & \textbf{84.93}   \\ \hline
    \end{tabular}
    \end{table*}
\end{center}

\begin{center}
\begin{table}[]
    \caption{The effect of our proposed InteractionModule (IM) and DilatedGroup (DG) which is evaluated on Figaro1k testset.\label{tab3}}
    \centering
    \begin{tabular}{ll|l}
    \hline
    \multicolumn{2}{c|}{Method} & \multicolumn{1}{c}{\multirow{2}{*}{mIoU}} \\
    IM           & DG           & \multicolumn{1}{c}{}                      \\ \hline
    -           & -            & 74.13                                     \\
    -           & \checkmark            & 75.67                                     \\
    \checkmark            & -           & 77.32                                     \\
    \checkmark           & \checkmark            & \textbf{78.39}                                     \\ \hline
    \end{tabular}
    \end{table}
\end{center}

\begin{center}
    \begin{table*}[h]%
        \caption{Results on CamVid test dataset. `-' indicates that the corrsponding result is not provided by the methods.\label{tab4}}
        \centering
        \begin{tabular}{l|cc}
        \hline
        Model        & \multicolumn{1}{l}{Time(ms)} & \multicolumn{1}{l}{mIoU(\%)} \\ \hline
        DPN\cite{Yu22018}          & 830                          & 60.1                         \\
        SegNet\cite{Badrinarayanan2017}       & 217                          & 46.4                         \\
        ICNet\cite{Zhao2018}        & 36                           & 67.1                         \\
        ENet\cite{Paszke2016}         & -                            & 51.3                         \\
        BiSeNet\cite{Yu2018}      & -                            & \textbf{68.7}                \\
        DFANet\cite{Li2019}       & 6                            & 59.3                         \\
        \textbf{HLNet} (ours) & \textbf{5.3}                 & 56.3                         \\ \hline
        \end{tabular}
    \end{table*}
\end{center}

\begin{center}
\begin{table*}[h]%
    \caption{Classification accuracy of different methods in different color spaces.\label{tab5}}
    \centering
    \begin{tabular}{l|ccc}
    \hline
                                  & \multicolumn{1}{l}{RGB} & \multicolumn{1}{l}{HSV} & \multicolumn{1}{l}{YCrCb} \\ \hline
    Histogram (8 bins)            & 75\%                    & 78\%                    & 73\%                      \\
    Histogram with PCA (256 bins) & 77\%                    & -                       & -                         \\
    Color Moment                  & 73\%                    & 77\%                    & \textbf{80\%}             \\ \hline
    \end{tabular}
    \end{table*}
\end{center}

\end{document}